\documentclass[runningheads]{llncs}
\usepackage{graphicx}
\usepackage{color, colortbl}
\usepackage{multirow}
\usepackage{mathtools}
\usepackage[table,xcdraw]{xcolor}
\usepackage{wrapfig}
\begin{document}
\title{Weakly Supervised Segmentation by A Deep Geodesic Prior}
\titlerunning{Weakly Supervised Segmentation by A Deep Geodesic Prior}
\author{Aliasghar Mortazi\inst{1,2,3,}\thanks{This work was done partially during internship at Boston Children's Hospital under the supervision of Dr. Kurugol and was supported partially by Crohn’s and Colitis Foundation of America’s (CCFA) Career Development Award and  AGA-Boston Scientific Technology and Innovation Award.
}, Naji Khosravan\inst{1}, Drew A. Torigian\inst{2}, Sila Kurugol\inst{3}, Ulas Bagci\inst{1}}
% index{Mortazi, Aliasghar}
% index{Khosravan, Naji}
% index{Torigian, Drew}
% index{Kurugol, Sila}
% index{Bagci, Ulas}

\authorrunning{A. Mortazi et al.}
\institute{Center for Research in Computer Vision (CRCV),\\ School of Computer Science, University of Central Florida, Orlando, FL, U.S.A.\and Medical Image Processing Group (MIPG),\\ Department of Radiology, University of Pennsylvania, Philadelphia, PA, U.S.A.\and Computational Radiology Laboratory(CRL), \\Department of Radiology, Boston Children's Hospital and Harvard Medical School, Boston, MA, U.S.A.}
%\email{}
%\url{} 
%\email{}
%
\maketitle              % typeset the header of the contribution
\vspace{-6mm}
\begin{abstract}
The performance of the state-of-the-art image segmentation methods heavily relies on the high-quality annotations, which are not easily affordable, particularly for medical data. To alleviate this limitation, in this study, we propose a weakly supervised image segmentation method based on \textit{a deep geodesic prior}. We hypothesize that integration of this prior information can reduce the adverse effects of weak labels in segmentation accuracy. Our proposed algorithm is based on a prior information, extracted from an auto-encoder, trained to map objects' geodesic maps to their corresponding binary maps. The obtained information is then used as an extra term in the loss function of the segmentor. In order to show efficacy of the proposed strategy, we have experimented segmentation of cardiac substructures with clean and two levels of noisy labels (L1, L2). Our experiments showed that the proposed algorithm boosted the performance of baseline deep learning-based segmentation for both clean and noisy labels by $4.4\%$, $4.6\%$(L1), and $6.3\%$(L2) in dice score, respectively. We also showed that the proposed method was more robust in the presence of high-level noise due to the existence of shape priors. %This observation supports the hypothesis that geodesic prior was more informative than conventional shape information.% ours is a more informative prior compared to previous methods, in the presence of noise. %The proposed algorithm is a significant step toward using~\textit{cheap} and~\textit{low quality} annotations to develop high performance automatic segmentation tools in general and medical imaging field in particular. 
%;~\textit{(2)} weak supervision allows to conduct the segmentation with weakly labeled data

\end{abstract}

\section{Introduction}
Driven by deep learning, artificial intelligence (AI) has attracted a widespread interest towards solving many challenging clinical problems. In medical AI applications, image segmentation is one of the mostly affected field from deep learning as it is often the first step for many image analysis tasks (shape analysis, volume measurements, and computer aided diagnosis). Since manual measurements are very expensive, time consuming, and  prone to inter- and intra-observer variations, having an automated, accurate, and efficient segmentation tool is the ultimate goal in many medical systems. In the deep learning era, numerous works have been published, showing feasibility of deep learning in segmentation of radiology images. However, most of these works focus on new network architectures adopted to the medical problem, and they rarely consider the fundamental challenge of the medical AI: availability of precisely annotated data.
%a continuous challenge in medical AI due to expensive and time consuming nature of the expert annotations. 

%To deal with noisy annotations, there have been  semi-supervised~\cite{khosravan2018semi,bai2017semi} and weekly-supervised~\cite{kervadec2018size,xu2014weakly} methods proposed, but mostly for diagnosis/classification tasks. Regarding image segmentation, ~\cite{weakse1} proposed a weakly supervised  method based on multiple clustered instance learning for histopathology image segmentation. In~\cite{weakse2}, authors segmented Crohn's disease tissues from abdominal MRI using weakly annotated samples. Despite such promising efforts, weakly supervised methods are not fully adopted in medical AI systems yet.
%Inaccurate annotations can be easily obtained from non-experts with a minimal training through settings such as mechanical turk.   

In this work, we propose a weakly-supervised segmentation method coupled with a deep geodesic prior to solve 3D medical image segmentation problem in a robust manner. This prior is mainly introduced to improve the performance of segmentation networks, more specifically when the annotations are noisy (i.e., not excellent). We argue that our proposed method is a significant step toward using inexpert and noisy annotations to train deep models for image segmentation without sacrificing the accuracy. The deep geodesic prior is specifically designed to put more attention in constructing accurate edges from weak labels. Although the proposed strategy is generic and can theoretically be applied to any medical image segmentation problem, herein we focus on cardiac MRI analysis due to its clinical significance and challenging nature of the MR imaging~\cite{who}.

\textbf{Related works:} One of the most successful deep-learning based segmentation methods is based on an encoder-decoder style architecture, called \textit{SegNet}, proposed by~\cite{segnet}. Among many other architectures, U-Net~\cite{unet} has became the most popular due to its properties such as efficient flow of low-level features through skip connections from decoder to encoder. To decrease the highly occupied parameter space of U-Net architecture(s), new network architectures were also presented: for instance, segmentation capsules (SegCaps)~\cite{capsules}, densely-connected network~\cite{densely} and Tiramisu~\cite{tiramisu} have shown drastic decrease in parameter space, while maintaining relatively good accuracy compared to baseline U-Nets. 

%maybe we dont need the following for now.
%Furthermore, they have been more attempts to solve the segmentation problem with using different approaches, i.e the left atrium and proximal pulmonary veins were segmented by using multi-view CNNs by~\cite{cardiacnet}. 

The literature for integrating shape priors into image segmentation is vast, mostly from pre-deep learning era. A mainstream approach is to construct a shape prior from a set of training samples represented implicitly by signed distance functions~\cite{kurugo,levelset}. In the deep learning era, Zotti et al.~\cite{cnnsp} used image registration to align shape priors and created atlas(es) to guide segmentation. Simply, authors have used this atlas for adding an extra loss term to the segmentation network. Modeling a prior (in shape or appearance) from medical images is still a challenging task due to highly diverse appearance, shape, and size of the anatomical objects. The first attempt to model shape prior with deep features was done by training an auto-encoder (AE) for creating features from the binary labels~\cite{acnn}. The AE was trained to reconstruct the binary input images in its output with a fully-connected layer as a bottleneck to capture the shape features. Then, these shape features were integrated into the segmentation network through an appropriate loss term. While the work in~\cite{acnn} is promising, it is not entirely clear whether the local anatomical variations are captured in detail.

We hypothesize that, if modeled correctly, prior information can lead to a more robust segmentation even when the labels are noisy (i.e., labels annotated by non-experts). To test this hypothesis, we propose a novel method for learning the prior from the geodesic maps of multiple objects. Then, an AE-like network is used to generate the original binary images from their corresponding geodesic maps. Finally, the features from the trained AE are used as a prior to be integrated into the segmentor for better guidance and performance improvement.

%\subsection{Cardiac Image Analysis}
%According to World Health Organization (WHO), cardiovascular disease (CVDs) are cause of 17.9 million death every year, which is 31\% of all deaths in the world~\cite{who}.One of the major indicator of predicting or diagnosis CVDs is measuring the Ejection Fraction (EF). 

%Usually , radiologists have to measure the volumes of the LV in two time points: end-diastole and end-systole which are time consuming and also they are prone to inter and intra-observer errors. So, several approaches beased on deep learning methods have been proposed to measure these volumes by segmenting LV and myo in end-systole and end-diastole in recent years. i.e Xue and et al. have proposed a deep-learning based method in~\cite{quan} for LV analysis. In~\cite{auto} a method to learn the segmentation architecture for segmenting LV and myo have been proposed. So, due to the importance of segmentation in cardiac image analysis, we have applied our proposed method to the ACDC 2017 dataset which include the annotation for LV, RV and myocardium of LV in cine-MR images.

\section{Method}
Our framework consists of two main components: (1) the segmentation network (or \textit{segmentor} in short), and (2) the geodesic prior learning network. While our segmentation network assigns a class label to each pixel of the input image, our second network (AE) learns a prior from geodesic maps, generated for each object of interest (for multiple objects). We anticipate (and show later in the results section) that incorporating a well-designed prior into the segmentation network improves its performance, especially in the presence of inaccurate labels. 

\begin{figure}[ht]
\centering
\vspace{-4mm}
\includegraphics[width=1\textwidth]{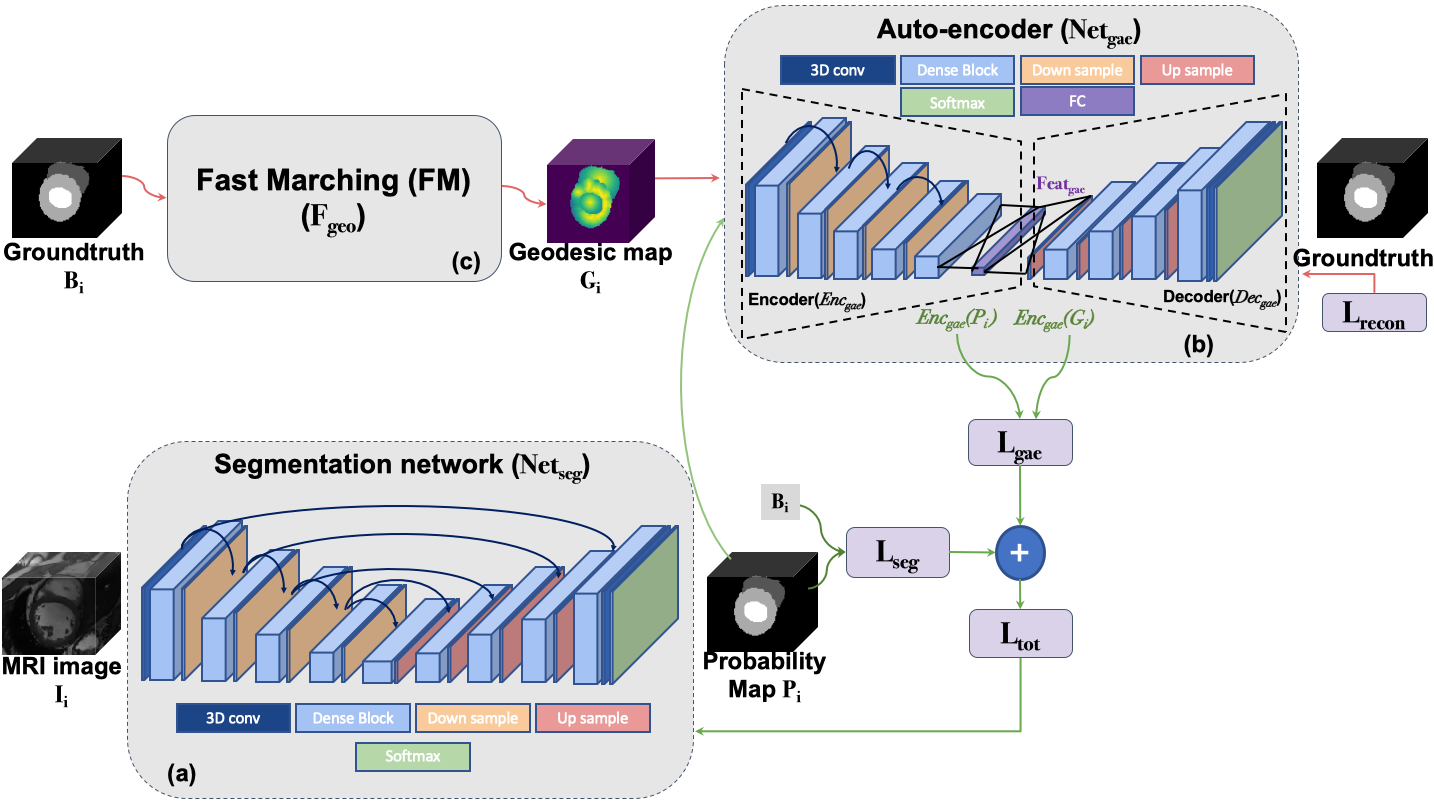}
\caption{The proposed framework has two main components: (1) segmentation network: assigns a class label to each pixel, and, (2)~\textit{AE} which learns a prior from geodesic maps. Green arrows show the flow of training of the segmentor and red arrows show the flow of training of GAE. Note that in test phase only segmentor is used. Also, the in our method the noisy annotations are used for whole training process.      
\label{fig:overview}}
\vspace{-5mm}
\end{figure}

The overview of our approach is illustrated in Figure~\ref{fig:overview}. The segmentation network ($Net_{seg}$) (Figure~\ref{fig:overview}(a)) is an encoder-decoder architecture with 3D kernel convolutions. We utilize skip connections in the form of dense connection throughout this network~\cite{densely}. For the prior learning network (AE), noisy annotations (or binary ground truths) are used to generate geodesic maps (Figure~\ref{fig:overview}(c)). Then, the geodesic auto-encoder (GAE) is designed to generate binary ground truths from geodesic maps. Once trained, GAE can be used to calculate two sets of bottleneck features: features resulted from feeding the geodesic map to GAE, and features resulted from feeding the corresponding $Net_{seg}$'s output probability map to GAE. Finally, the distance between these two feature vectors are used to form an extra term in the loss function of the $Net_{seg}$ (Figure~\ref{fig:overview}).

We define the segmentation network as a function, mapping a gray-scale 3D input image $I_i(I_i \in R^3)$ into a probability map $P_i$, ($i\in\{1, 2, \dots, N\}$), $N$ being number of 3D images:

\begin{equation}
    P_{i} = Net_{seg}(I_i, \theta_{seg}),
\end{equation}
\noindent where $\theta_{seg}$ are the parameters of $Net_{Seg}$, trained to minimize $\mathcal{L}_{tot}$ defined in Equation~\ref{eq:l_tot}. A geodesic map, on the other hand, is generated from ground truth binary images $B_i$ as $G_i = F_{geo}(B_i)$ and then a GAE network ($Net_{gae}$) is trained to generate binary image from this corresponding geodesic map (explained in Section \ref{sec:geo}). The GAE consists of an encoder, a fully connected (FC) layer, and a decoder. The encoder and FC layers are the feature extraction ($Enc_{gae}$) parts, mapping the input geodesic map to the feature vector $Feat_{gae}=Enc_{gae}(G_i)$ of length $L_{feat}$. The decoder ($Dec_{gae}$) reconstructs the corresponding binary image(s) from the $Feat_{gae}$. Hence, the geodesic network can be formulated as: 

\vspace{-2mm}
\begin{equation}
    \hat{B}_i = Net_{gae}(G_i, \theta_{gae}) = Dec_{gae}(Enc_{gae}(G_i, \theta_{enc}), \theta_{dec}), 
\end{equation}
\noindent where $\theta_{gae}$ are the parameters of $Net_{gae}$, trained to minimize the binary map reconstruction loss $\mathcal{L}_{recon}$ in $Net_{gae}$. $\mathcal{L}_{recon}$ is a cross-entropy loss between ground-truth and $Net_{gae}$'s output and $\theta_{gae}=\theta_{enc}\cup \theta_{dec}$. Since $Net_{gae}$ is designed to learn the relation between geodesic maps and their corresponding binary maps, $Feat_{gae}$ contains high-level features inferred from shapes and texture of the objects of interests. This encoded knowledge can be used as an extra term of supervision for better training of $Net_{seg}$. For each training sample $i$, we calculate a loss function $\mathcal{L}_{gae}(Enc_{gae}(P_i), Enc_{gae}(G_i))$ to be back-propagated into the segmentation network along with loss of the segmentor itself. The total loss function of the segmentator is then represented as:

\begin{equation}
    \mathcal{L}_{tot}=\sum_i \mathcal{L}_{seg}(P_i, B_i)+\mathcal{L}_{gae}(Enc_{gae}(P_i), Enc_{gae}(G_i)).
\end{equation}\label{eq:l_tot}
We first train $\theta_{gae}$ with $\mathcal{L}_{recon}$. Once trained, $\theta_{seg}$ is updated with the loss function $\mathcal{L}_{tot}$, while $\theta_{gae}$ are fixed.   
\vspace{-3mm}
\subsection{Network Architecture for Segmentation}\label{sec:arch}
\begin{wrapfigure}{rt}{0.43\textwidth}
\vspace{-5mm}
\centering
\includegraphics[width=.45\textwidth, height=1.2cm]{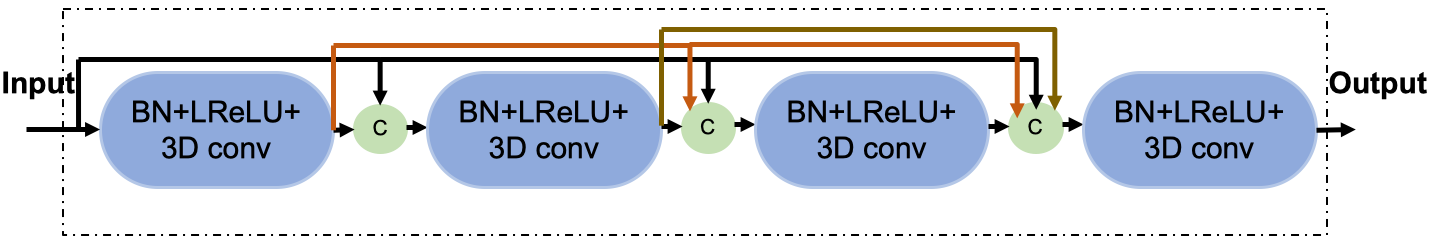}
\caption{Dense Block (DB) content.~\textit{C}: concatenation operation,~\textit{BN}: Batch normalization,~\textit{LReLu}: Leaky ReLu activation, and~\textit{3D conv}: 3D convolution with a $3 \times 3 \times 3$ filter.
\label{fig:dense_block}}
\vspace{-5mm}
\end{wrapfigure}
We extend fully convolutional dense nets, called ~\textit{Tiramisu}~\cite{tiramisu}, from 2D to fully 3D. The details of the adapted network are shown in Figure~\ref{fig:overview}(a). The encoder and decoder include four dense blocks, each. Within dense blocks, there are four 3D convolution layers followed by a \textit{Batch Normalization} (BN) layer with \textit{Leaky ReLu} nonlinear activation. The size of the convolution kernels are set to $3 \times 3 \times 3$ in all convolution layers (Figure~\ref{fig:dense_block}). The number of filters in the first convolution layer and the~\textit{growth rate} are set to~\textit{16} for an optimal performance after extensive explorations. The number of output filters in a dense block is $N_f(\mathbf{X}_{\it{l}})=N_f(\displaystyle \underset{{\it{l}'=0}}{\overset{{\it{l}'=\it{l}-1}}{\|}} N_f(\mathbf{X}_{\it{l}'}))$, where $\it{l}=\{1,2,\dots,\it{L}\}$ and $\|$ is the concatenation. 
%For initialization $N_f(\mathbf{X}_{\it{-1}})$ and $N_f(\mathbf{X}_{\it{0}})$ are considered as~\textit{empty cell} and $\mathbf{X}_{\it{1}}$ (for first block $\mathbf{X}_{\it{1}}=$~\textit{growth rate}), respectively, and there are $\it{L}$ layers inside each block.

%\begin{figure}[ht]
%\centering
%\includegraphics[width=1\textwidth]{seg_arch.png}
%\caption{Architecture of segmentation network 
%\label{fig:seg_arch}}
%\end{figure}

The encoder includes four 3D max pooling operations as transition layers after each dense block. The pooling operation is set to downsample the input size by 2 in $x$-$y$ plane. Downsampling is not applied to $z$ direction due to its low resolution (but could be applied for other settings). Similarly, the decoder includes four up-samplers (using bilinear interpolation) as transition layers. Each up-sampler is designed to double the size of its input. Finally, the last layer contains a convolution layer following by a \textit{softmax} function to introduce a notion of probability map in the output. We use~\textit{Adam} optimizer to minimize the $\mathcal{L}_{tot}$ in $Net_{seg}$. $\mathcal{L}_{seg}$ and $\mathcal{L}_{gae}$ are designed with~\textit{Cross Entropy} and~\textit{Mean Square Error} functions, respectively.

%This segmentation network in our framework is responsible for assigning a class probability value to each pixel in the input. In the next section we describe the details of our prior learning network and how it will contribute to the segmentation.
\vspace{-3mm}
\subsection{Learning a Geodesic Prior} \label{sec:geo}
Most existing literature related to prior incorporation into segmentation utilize \textit{accurate} binary labels for extracting shape information~\cite{acnn,cnnsp}. Unlike the mainstream studies, we propose to use geodesic maps to increase robustness of the priors when dealing with \textit{noisy} labels, which has never been done before. This approach can particularly be beneficial when the object has complex boundary information to be delineated. In this study, geodesic maps are generated from labels, regardless of being noisy or clean. We expect the proposed geodesic map to capture more information then conventional shape priors.% and lead into aOur specific design leaded to a more informative information extraction through our GAE, which leads to more robustness in the presence of noise.

For each object in our images, we compute an independent geodesic distance map from its binary map $B_i$ by using the Fast Marching (FM) approach~\cite{fastmarch}. FM is a numerical method to solve boundary value problems of the Eikonal as:

\begin{equation}
%F_{geo}(x)=
\begin{aligned}
\begin{dcases}
    F(x) |\Delta T(x)| = 1 ,& \forall x \notin S,\\
    F(x)=0,              & \forall x \in S,
\end{dcases}
   \end{aligned}
   \label{equ:distance}
\end{equation}

\noindent describing the evolution of a contour as a function of time, $T(x)$, with the speed of $F(x)$ in the normal direction at a point $x$ on the propagating surface starting from the zero-level $S$. With a specified speed, $F(x)$, the time when the contour crosses point $x$ can be computed by solving equation~\ref{equ:distance}. In this setting, the special case of $F(x)=1$ gives the signed distance of every point $x$ from $S$. In our case, since we have multiple objects (i.e., $3$ objects: LV, RV, Myocardium), we defined $S$ as the center of mass of the all closed objects. In our experimental setup, we have also an object with non-Jordan surface (i.e., Myocardium, having donuts shape). In order to include such objects in the geodesic computation, we simply define the the skeleton of the shape and the distances of all the points within each object are computed from their zero line contours $S$. These maps (obtained from each object) are combined in $n$-channels (i.e., 3 in our experiments: LV, RV, Mayo) and fed into the auto-encoder as described below.

The proposed AE architecture ($Net_{gae}$), for learning prior information, is illustrated in Figure~\ref{fig:overview}(b). This architecture is very similar to the segmentor with a FC layer in the middle (instead of a convolution layer) to generate deep geodesic features. Also, in order to increase the robustness of these features, there is no skip connections from encoder to decoder. Both encoder and decoders include four DBs and the filters size in each convolution layer was set to $3 \times 3 \times 3$. The growth rate for the encoder part is set to $16$ (empirically) as in the segmentor and \textit{Adam} optimizer is used to minimize $\mathcal{L}_{recons}$ (~\textit{Cross Entropy} loss).

\vspace{-3mm}
\section{Experiments and Results}
\vspace{-2mm}
To show the robustness of our algorithm and its performance on noisy labels, we ran all the experiments on both expert as well as two levels of noise in the labels (L1 and L2). We reported Dice Index (DI) and Hausdorff distance (HD) in Table~\ref{tab:results}. Also, in cases where we were dealing with noisy labels, there was an upper bound for the performance of the networks. This upper bound was due to lack of information in the presence of noise. To have a sense of this upper bound, for the sake of a more extensive and fair comparison, DI and HD of generated noisy labels with respect to clean ground truths are reported in this table (~\textit{Upper boundary} columns). Higher DI and lower HD indicate a superior segmentation performance. While training of the networks were done using weak/noisy labels, validation was performed on expert/accurate labels.
%It is also noteworthy for understanding the power of all methods in learning from inexpert labels, the evaluation (in both validation and test data) have been done on the expert labels.

\textbf{Data set:} We used the cine MR cardiac data set from Automated Cardiac Diagnosis Challenge (ACDC) MICCAI challenge 2017~\cite{acdc}. The images in this data set were obtained from two MRI scanners of different magnetic strengths (1.5T and 3.0T). Cine MR images were acquired in breath-hold with a retrospective or prospective gating and with a SSFP sequence which LV was covered by sries of short axis slices with a thickness of 5 mm. The spatial resolution of the images goes from 1.37 to 1.68 $mm^2$/pixel and 28 to 40 images cover completely or partially the cardiac cycle. Out of 150 cine MR images, we used 100 for training and validation (including expert annotation for LV, Myo, and RV at ES and ED), and the remaining 50 for testing (with online evaluation). Subject categories (5) are the following: 30 normal subjects, 30 with myo infarction, 30 with dilated cardiomyopathy, 30 with hypertrophic cardiomyopathy, and 30 with abnormal right ventricle which make the segmentation task challenging over different cases. 
\begin{wrapfigure}{rt}{0.43\textwidth}
\vspace{-5mm}
\centering
\includegraphics[width=0.45\textwidth, height=1.8cm]{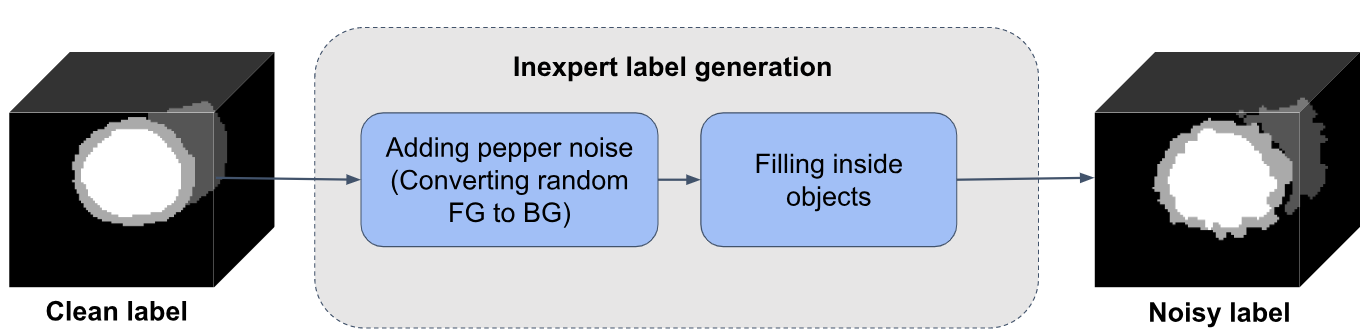}
\caption{Generating noisy labels. Binary images went through a two-step process of adding pepper noise and filling inside object which makes the boundaries inaccurate.
\label{fig:noisy}}
\vspace{-5mm}
\end{wrapfigure}

\textbf{Generating noisy labels:} The current annotations at ES and ED in ACDC dataset are considered as expert annotations (as clearly defined by the challenge organizers). Usually, the inexpert annotations include some under-segmentation and/or over-segmentation. This is due to lack of naive annotator's knowledge in finding the edges. Thus, in order to mimic such inexpert annotations (weak labels), we manipulated the ground truths as follows: first we obtained the outer shell of each object by applying~\textit{erosion} to the binary image and then calculate the difference between the original binary image and eroded one. Then, salt-pepper noises were added to each object's binary shell randomly and~\textit{filling} was applied to the shell. This process effected \textit{only} edges of the objects without changing the background, resulting in a shape with distorted boundaries. Finally, eroded binary edges was added to the distorted shell. A sample of weak labels vs. expert labels is shown in Figure~\ref{fig:noisy}. Participating radiologists confirmed the weak labels through visual evaluations. %It is noteworthy that it is better idea to get inexpert annotations from naive people (i.e Amazon Mechanical Turk), but the regulations from ACDC challenge didn't allow to get such annotations. 

\textbf{Baseline models for comparisons:} We have conducted several baseline architectures in order to show the strength of our proposed method. First, the segmentor was trained only with $\mathcal{L}_{seg}$ and without using prior information with both of the inexpert and expert annotations. Also, in order to illustrate the advantage of using the geodesic maps instead of binary maps in modeling shape information, the segmentation network was trained with the prior information obtained from binary AE (\textit{segmentation + binary labels shape}). In this baseline, the AE was trained to reconstruct binary map in its output from its corresponding binary input map (instead of geodesic in our method). The results of this baseline (\textit{Binary Prior}) and our proposed method (\textit{Geodesic Prior}) for inexpert and expert annotations are reported in Table 1.

% \usepackage{multirow}
\vspace{-4mm}
\begin{table}[h]
\caption{DI and HD are reported for both expert and inexpert labels. \label{tab:results}}
\vspace{-2mm}
\resizebox{\columnwidth}{!}{
\begin{tabular}{|c|c|
>{\columncolor[HTML]{FFFFFF}}c |
>{\columncolor[HTML]{FFFFFF}}c |
>{\columncolor[HTML]{FFFFFF}}c |
>{\columncolor[HTML]{FFFFFF}}c |
>{\columncolor[HTML]{FFFFFF}}c |
>{\columncolor[HTML]{FFFFFF}}c |
>{\columncolor[HTML]{FFFFFF}}c |
>{\columncolor[HTML]{FFFFFF}}c |
>{\columncolor[HTML]{FFFFFF}}c |
>{\columncolor[HTML]{FFFFFF}}c |
>{\columncolor[HTML]{FFFFFF}}c |}
\hline
\multicolumn{2}{|c|}{\cellcolor[HTML]{C0C0C0}Labels} & \multicolumn{3}{c|}{\cellcolor[HTML]{C0C0C0}Expert Labels} & \multicolumn{4}{c|}{\cellcolor[HTML]{C0C0C0}Inexpert Labels(L1)} & \multicolumn{4}{c|}{\cellcolor[HTML]{C0C0C0}Inexpert Labels(L2)} \\ \hline
\multicolumn{2}{|c|}{\cellcolor[HTML]{C0C0C0}Methods} & \cellcolor[HTML]{C0C0C0}\begin{tabular}[c]{@{}c@{}}Seg. \\ Net.\end{tabular} & \cellcolor[HTML]{C0C0C0}\begin{tabular}[c]{@{}c@{}}Binary \\ Prior\end{tabular} & \cellcolor[HTML]{C0C0C0}\begin{tabular}[c]{@{}c@{}}Geodesic \\ Prior\end{tabular} & \cellcolor[HTML]{C0C0C0}\begin{tabular}[c]{@{}c@{}}Seg. \\ Net.\end{tabular} & \cellcolor[HTML]{C0C0C0}\begin{tabular}[c]{@{}c@{}}Binary \\ Prior\end{tabular} & \cellcolor[HTML]{C0C0C0}\begin{tabular}[c]{@{}c@{}}Geodesic \\ Prior\end{tabular} & \cellcolor[HTML]{C0C0C0}\begin{tabular}[c]{@{}c@{}}Upper \\ boundary\end{tabular} & \cellcolor[HTML]{C0C0C0}\begin{tabular}[c]{@{}c@{}}Seg. \\ Net.\end{tabular} & \cellcolor[HTML]{C0C0C0}\begin{tabular}[c]{@{}c@{}}Binary \\ Prior\end{tabular} & \cellcolor[HTML]{C0C0C0}\begin{tabular}[c]{@{}c@{}}Geodesic \\ Prior\end{tabular} & \cellcolor[HTML]{C0C0C0}\begin{tabular}[c]{@{}c@{}}Upper \\ boundary\end{tabular} \\ \hline
 & LV & 0.854 & 0.879 & \textbf{0.885} & {\color[HTML]{000000} 0.831} & {\color[HTML]{000000} 0.867} & {\color[HTML]{000000} \textbf{0.878}} & {\color[HTML]{000000} 0.880} & 0.811 & 0.856 & \textbf{0.873} & 0.869 \\ \cline{2-13} 
 & RV & 0.803 & \textbf{0.851} & 0.847 & {\color[HTML]{000000} 0.791} & {\color[HTML]{000000} 0.828} & {\color[HTML]{000000} \textbf{0.836}} & {\color[HTML]{000000} 0.835} & 0.762 & 0.811 & \textbf{0.831} & 0.824 \\ \cline{2-13} 
 & MYO & 0.771 & 0.816 & \textbf{0.826} & {\color[HTML]{000000} 0.762} & {\color[HTML]{000000} 0.798} & {\color[HTML]{000000} \textbf{0.810}} & {\color[HTML]{000000} 0.812} & 0.751 & 0.780 & \textbf{0.809} & 0.801 \\ \cline{2-13} 
\multirow{-4}{*}{DI} & Ave. & 0.809 & 0.849 & \textbf{0.853} & {\color[HTML]{000000} 0.795} & {\color[HTML]{000000} 0.831} & {\color[HTML]{000000} \textbf{0.841}} & {\color[HTML]{000000} 0.842} & 0.775 & 0.816 & \textbf{0.838} & 0.831 \\ \hline
 & LV & 14.79 & \textbf{10.08} & 10.14 & {\color[HTML]{000000} 15.87} & {\color[HTML]{000000} 12.57} & {\color[HTML]{000000} \textbf{11.73}} & {\color[HTML]{000000} 11.75} & 14.89 & 13.03 & \textbf{11.65} & 11.81 \\ \cline{2-13} 
 & RV & 17.77 & 13.77 & \textbf{13.45} & {\color[HTML]{000000} 18.89} & {\color[HTML]{000000} 17.01} & {\color[HTML]{000000} \textbf{15.14}} & {\color[HTML]{000000} 15.15} & 17.57 & 17.53 & \textbf{15.44} & 15.76 \\ \cline{2-13} 
 & MYO & 16.53 & 12.58 & \textbf{12.04} & {\color[HTML]{000000} 16.91} & {\color[HTML]{000000} 13.95} & {\color[HTML]{000000} \textbf{12.73}} & {\color[HTML]{000000} 12.70} & 16.47 & 15.12 & \textbf{12.88} & 13.12 \\ \cline{2-13} 
\multirow{-4}{*}{\begin{tabular}[c]{@{}c@{}}HD\\ (mm)\end{tabular}} & Ave. & 16.36 & 12.14 & \textbf{11.88} & {\color[HTML]{000000} 17.22} & {\color[HTML]{000000} 14.51} & {\color[HTML]{000000} \textbf{13.20}} & {\color[HTML]{000000} 13.20} & 15.64 & 15.23 & \textbf{13.32} & 13.56 \\ \hline
\end{tabular}
}
\end{table}
\vspace{-5mm}

\textbf{Implementation Details:} As a pre-processing step, we applied the anisotropic filtering to reduce noise from MRI, histogram matching to standardize MRI intensities, and all images were resized to $200\times200\times10$ by using B-spline interpolation. First the~\textit{GAE} was trained (early-stopping) and then the deep geodesic features ($Feat_{gae}$) were extracted from training data. Then, during training of $Net_{seg}$ the output probability maps of the $Net_{seg}$ were passed though $Enc_{gae}$ and then the loss ($\mathcal{L}_{gae}$) between two feature vector was calculated and back-propagated though the $Net_{seg}$. Finally, Conditional Random Field is used for post-processing. we used 80 MR images for training, the 20 images were used as validation, and the 50 images were used for test (with online evaluation). Also, we have used NVIDIA Quadro P6000 GPU for training the networks.
\vspace{-2mm}
\section{Conclusion}
\vspace{-2mm}
In this study, we propose a novel framework incorporating a deep geodesic prior information into the segmentation framework. Our AE network is capable of learning high-level features from generated geodesic maps for multiple objects. We show that our proposed approach outperforms the state-of-the-art methods both on clean and noisy labels with several key advantages. First, incorporating prior information improves segmentation results even with imperfect ground truths. Second, more specifically, shape prior is shown to be useful both in Euclidean and Geodesic distance based evaluations, and geodesic priors are shown to be more accurate than the former. Furthermore, the proposed network is capable of performing fully 3D image segmentation unlike most 2D methods in the literature, and can handle multiple objects too. One may argue to obtain inexpert annotations directly from inexpert annotators for more realistic evaluations, but the regulations for medical imaging data sharing in general (and ACDC challenge in particular) didn't allow to get such annotations. Our future work will comprehensively test our hypothesis for different components of our present study such as different clinical imaging problem, large number of inexpert annotators, and inspecting the results based on object type and size.

%Last, but not least, we believe that our work  is a significant step toward accurate segmentation even the labels are weak, which is the case in many clinical imaging problems. %The use of geodesic priors for multi-object segmentation in the deep learning setting as well as its integration with weakly supervised learning scheme are all new, and never been done before.
\vspace{-3mm}
\bibliographystyle{splncs04}
\bibliography{mybibfile}

\end{document}